# Corporate Language Model (CLM): Transforming Tacit and Fragmented Enterprise Knowledge into a Sovereign, Auditable, and Executable Corporate Intelligence Layer

**Fabricio Colvero Avini**
*Business Graduate Program – Unisinos University*
*Care Intelligence, Porto Alegre, Brazil*
fabricio@careintelligence.com.br

**Guilherme Trez**
*Business Graduate Program – Unisinos University*
*Care Intelligence, Porto Alegre, Brazil*
gtrez@unisinos.br

## Abstract

Enterprise AI deployments increasingly fail not because of model inadequacy, but because organizations lack a structured substrate to encode how they actually decide, negotiate, and execute. Generic Large Language Models (LLMs) carry no priors about a firm's idiosyncratic ontology, tacit conventions, or governance constraints; Retrieval-Augmented Generation (RAG) approaches remain brittle and offer no path from recommendation to executable action; static playbooks encode logic but cannot reason or adapt. These constraints motivate a foundation-centric architecture in which tacit-knowledge capture, ontological grounding, sovereign deployment, and auditable actuation are treated as co-designed, non-negotiable constraints from the outset rather than as features added later.

This paper introduces the Corporate Language Model (CLM), an architectural framework that transforms a firm's structured, unstructured, multimodal, and tacit knowledge into a unified, ontology-grounded enterprise knowledge foundation on top of which reasoning, agents, and governed execution are composed. CLM is organized as five interlocking capability planes - Foundation, Tactical, Wisdom-Out, Governance, and a transversal Deep Security Layer - and rests on four architectural pillars: a Neurosymbolic Mesh coupling generative models with an enterprise knowledge graph; a Skill Graph in which reusable tactics, personas, objections, and goals are typed and composed; Living Digital Twins modeling functional areas as reasoning surrogates; and a Deep Security Layer enforcing sovereignty, traceability, and human oversight as architectural floors. A Spec-as-Code paradigm closes the gap between grounded intent and executable artifact, producing mutually consistent business specifications, technical specifications, behavior scenarios, and interface contracts from a single governed derivation chain. CLM is one principled instantiation of this foundation-centric class, oriented toward neurosymbolic coupling and governance-by-design.

The paper makes four contributions. It defines CLM as a conceptual architecture organized around five capability planes and four architectural pillars, establishing the framework as a distinct object of study in the enterprise AI literature. It introduces the Skill Graph formalism as a mechanism for compositional explainability by construction - where the reasoning path is the explanation, requiring no post-hoc approximation - and provides a compact formal core. It proposes the Wisdom Listener effect as a theoretical property by which tacit-capable foundations compound in value with organizational use, connecting this property to established mechanisms in dynamic capabilities and organizational learning. Early operational evidence from a production deployment in a regulated healthcare environment - a JCI-accredited tertiary hospital in Brazil - instantiates three of the six cumulative maturity stages and demonstrates the sovereignty and auditability constraints under Brazil's LGPD framework.

The central contribution is a repositioning of the enterprise AI design question: from *which model performs best* to *which foundation is the deepest*. CLM makes the conceptual case that tacit knowledge, properly captured and governed, constitutes a firm's most durable executable asset.

## 1 Introduction

The dominant question in enterprise AI has shifted. Organizations that moved past the feasibility stage - can we use LLMs? - now confront a more consequential one: why does our AI keep failing on our own business? The failure is not linguistic. State-of-the-art language models produce fluent, contextually coherent outputs. The failure is epistemic: these models carry no priors about how a specific firm decides, what exceptions its contracts encode, how its veterans actually close deals, or which escalation paths its controllers recognize as legitimate. That knowledge exists - it is operationally decisive - but it lives in chat threads, voice notes, meeting fragments, and the margins of playbook PDFs that no process ever formalized. It is tacit, fragmented, and effectively invisible to both human newcomers and to the generic models being deployed at scale. The pattern is most visible in regulated environments such as the Brazilian hospital setting in which the architecture described here was first deployed, where every clinical workflow carries its own exceptions and where the cost of an unexplainable recommendation is measured in accreditation and patient safety rather than in user satisfaction.

Three classes of response have emerged, and each is structurally insufficient. Retrieval-Augmented Generation (RAG) grounds model outputs in a private corpus, but inherits the quality of that corpus, remains loosely coupled to the firm's ontology, and offers no mechanism to turn a recommendation into an executable action inside a legacy-heavy environment. Prompt engineering over a frontier model compensates for missing context at the margins but cannot substitute for the firm's accumulated reasoning patterns. Static playbooks - structured rule books, decision trees, scripted escalation paths - encode organizational logic faithfully but cannot adapt to context, cannot handle the edge cases that define operational reality, and decay the instant the business evolves faster than the document.

These are not isolated engineering shortcomings. They reflect five interlocking failure modes that recur across enterprise AI adoption regardless of industry or scale: tacit knowledge that is structurally undocumented; generic models with no firm-specific ontological priors; recommendations that carry no auditable logical path; data sovereignty constraints that most cloud-first architectures cannot satisfy; and the forced choice between rigid rule books and opaque agentic systems. Existing approaches treat these failure modes as features to be added later. This paper argues they must be treated as design constraints from the outset - and that this reordering is itself the central architectural insight.

The thesis of this paper is that closing this gap requires a different object of study: not a model, not a pipeline, but a foundation. We introduce the Corporate Language Model (CLM), an architecture that encodes a firm's knowledge, reasoning patterns, governance constraints, and execution contracts as a single, composable, auditable substrate. We are explicit about the logical status of the claim: the five failure modes motivate a class of foundation-centric architectures, and CLM is a specific neurosymbolic and governance-oriented instantiation of that class. A CLM is not a replacement for a generic LLM. It is the structured layer around it, below it, and at times orthogonal to it - the layer that transforms general linguistic competence into corporate executable intelligence.

This is, by design, a position paper: it advances architectural propositions and design commitments, situates them against related work, and states explicit, falsifiable criteria for the empirical validation that remains future work. It makes four contributions. First, it defines CLM as a conceptual architecture organized around five interlocking capability planes and four architectural pillars, establishing the framework as a distinct object of study in the enterprise AI literature. Second, it introduces the Skill Graph formalism - a typed, directed graph of reusable tactical reasoning units - as a mechanism for compositional explainability by construction, and gives it a compact formal core (Section 5.2). Third, it proposes the Wisdom Listener effect as a theoretical property by which tacit-capable foundations compound in value with organizational use. Fourth, it presents early operational evidence from a regulated healthcare deployment that instantiates three of the six cumulative maturity stages under real sovereignty and auditability constraints.

Two terminological notes before proceeding. First, CLM is used exclusively throughout this paper to denote Corporate Language Model - a usage orthogonal to its appearance in natural language processing (where CLM denotes causal language models) and in enterprise legal and contract management software (where CLM denotes contract lifecycle management systems). Second, the phrase "Enterprise Knowledge Foundation" is used descriptively to designate the organizational knowledge substrate that CLM produces and maintains: the accumulated, ontologically grounded, and queryable representation of a firm's explicit and tacit knowledge. We treat CLM as the architectural system and enterprise knowledge foundation as the substrate it constitutes.

The remainder of the paper is organized as follows. Section 2 formalizes the five failure modes as design constraints. Section 3 positions CLM against nine adjacent research streams. Sections 4 through 6 develop the architecture: the five capability planes, the four pillars, and the Spec-as-Code paradigm. Section 7 presents early empirical evidence and its explicit limitations. Section 8 identifies the strategic differentiators, Section 9 presents the discussion and research agenda, and Section 10 concludes.

## 2 Problem Statement

Five interlocking failure modes dominate enterprise AI adoption at scale. They are not independent: each amplifies the others, and together they define the design space that CLM is built to occupy. We treat them here as constraints on the architecture, not as features to be satisfied later. This ordering - constraints first, capability second - is itself part of the thesis.

***(P1) Tacit knowledge is the majority of what makes organizations effective, and it is structurally undocumented.***

The knowledge that determines how a firm actually operates - how its senior negotiators close deals, how its controllers decide what to approve, which client exceptions its contracts silently encode - is overwhelmingly tacit rather than explicit. Nonaka and Takeuchi [17] established that organizational knowledge creation depends on the continuous conversion between tacit and explicit forms; the failure to manage this conversion systematically produces organizations that cannot transfer, replicate, or scale their own competence. Davenport and Prusak [18] extended this observation to the information systems context, documenting how the most operationally decisive knowledge resists formalization precisely because it is embedded in practice, judgment, and exception rather than in procedure. In contemporary enterprise environments, this substrate lives in Slack threads, WhatsApp voice notes, recorded calls, the margins of proposal PDFs, and the gap between what the CRM registers and what actually happened. A system that cannot capture, structure, and preserve this tacit substrate cannot reason as the organization reasons - regardless of the linguistic capability of the underlying model.

***(P2) Generic large language models carry no firm-specific ontological priors.***

Foundation models are trained on corpora of general breadth and carry strong linguistic priors, but effectively no priors about a specific firm's idiosyncratic ontology, incentive structure, accumulated exceptions, or internal terminology. The difference between how a given firm defines a "qualified opportunity" and how the same term appears in general business text is not a marginal discrepancy - it is the entire operational content of the concept. Prompt engineering compensates for this gap only at the margins: it can inject context into a single interaction but cannot internalize the firm's reasoning patterns across sessions, domains, and edge cases. Fine-tuning on proprietary corpora is a partial remedy, but without ontological grounding, a fine-tuned model inherits the biases and gaps of its training corpus rather than the structured reasoning of the organization [7]. Görgen et al. [25] provide empirical confirmation through a case study of LLM deployment in enterprise modeling, finding that the more domain-specific the required knowledge, the less suitable generic LLMs prove to be - a result that holds regardless of model scale and that motivates the ontological grounding mechanism of CLM's Foundation Plane.

***(P3) Recommendations without auditable logical paths are operationally dangerous.***

For any decision that touches regulated processes, financial exposure, or customer-facing commitments, a recommendation without a traceable logical path is either unusable - legal departments reject it - or reckless - someone approves it without understanding the basis. This is not a niche concern: the NIST AI Risk Management Framework [12] identifies auditability and explainability as foundational requirements for responsible AI deployment in consequential domains, and the OWASP Top 10 for LLM Applications [11] documents how the absence of output traceability is among the most exploited vulnerabilities in production deployments. The inability to reconstruct why a model produced a given output is not merely a technical limitation - it is a governance failure that disqualifies the system from regulated contexts before it reaches production.

***(P4) Data sovereignty and regulatory constraints are structural, not discretionary.***

Large segments of the economy - healthcare, banking, insurance, and the public sector - operate under regulatory frameworks that prohibit sending primary operational data to external inference endpoints. In Brazil, the Lei Geral de Proteção de Dados (LGPD, Law 13,709/2018) makes this constraint legally binding rather than organizationally discretionary, and comparable frameworks - GDPR in Europe, HIPAA in the United States, and sector-specific analogues across most regulated jurisdictions - impose equivalent perimeter requirements. An architecture that assumes cloud-only frontier inference over firm data is structurally incompatible with these environments, not merely suboptimal. The maturation of compact, locally deployable language models [5] has made sovereign deployment technically viable; the architectural question is no longer whether sovereign inference is possible, but whether the surrounding framework is designed to exploit it.

***(P5) Static playbooks cannot adapt; live agentic systems cannot be audited.***

Organizations that recognize the limitations of generic models typically reach for one of two compensating mechanisms, and both fail at scale. Static playbooks encode organizational logic faithfully but cannot adapt to context, cannot handle the edge cases that define operational reality, and decay the instant the business evolves faster than the document. Agentic systems, by contrast, can reason and adapt, but their decision paths are opaque: the sequence of tool invocations, retrieval operations, and generation steps that produced a given output is typically unrecoverable after the fact. Wang et al. [16] survey the current landscape of LLM-based autonomous agents and identify auditability as a first-order unsolved problem across the field. Neither option is acceptable at enterprise scale: the first sacrifices adaptability, the second sacrifices accountability, and the gap between them is precisely where CLM is designed to operate.

CLM treats these five failure modes as non-negotiable constraints on architecture rather than as a prioritized backlog of features. They do not uniquely imply CLM; they imply a foundation-centric class of architectures, and the remainder of the paper argues that CLM is a principled and especially complete response within that class. The design implications of this reordering - constraints first, capability second - run through every section that follows.

## 3 Related Work

The nine streams reviewed in this section were identified through structured search of the Scopus, IEEE Xplore, ACM Digital Library, and arXiv databases, using the following terms and their combinations: *enterprise AI*, *tacit knowledge capture*, *retrieval-augmented generation*, *neurosymbolic AI*, *knowledge graph*, *digital twin*, *agentic systems*, *explainable AI*, and *domain-specific language models*. Selection was governed by direct relevance to the five design constraints established in Section 2: streams were included if they address at least one of P1 through P5, either as a partial solution whose limitations CLM extends, or as an established theoretical lineage on which CLM builds. The review covers foundational and recent work; for active areas such as RAG variants and neurosymbolic architectures, coverage is weighted toward publications from 2023 onward to reflect the current state of practice.

CLM stands at the intersection of these nine streams. We describe each and state, in the positioning paragraph that closes each subsection, precisely what CLM inherits, what it extends, and where it departs.

### 3.1 Organizational Knowledge Management

The foundational argument of this paper - that tacit knowledge is the majority of what makes organizations effective and is structurally underdocumented - rests on a body of research that predates AI entirely. Nonaka and Takeuchi [17] established the SECI model of knowledge creation, demonstrating that organizational effectiveness depends on the continuous conversion between tacit and explicit knowledge through socialization, externalization, combination, and internalization. Davenport and Prusak [18] extended this to information systems contexts, showing that the most operationally decisive knowledge resists formalization precisely because it is embedded in practice and exception rather than in procedure. What these foundational works lacked was a computational substrate capable of capturing, preserving, and reasoning over tacit knowledge at scale. CLM treats this gap as its primary design constraint: the Foundation Plane and tacit-capture mechanisms are, in architectural terms, the computational operationalization of the externalization and combination steps in the SECI model.

### 3.2 Retrieval-Augmented Generation and Its Corrective Variants

Classical RAG couples a retriever with a generator to ground LLM outputs in an external corpus. More recent work extends this in several directions: Self-RAG [2] trains a model to decide adaptively when to retrieve and to critique its own outputs via reflection tokens; Corrective RAG [3] adds a retrieval-quality evaluator that triggers corrective retrieval when quality is low; Hypothetical Document Embeddings [4] pivot retrieval through a synthetically generated document whose embedding navigates the corpus more robustly in zero-shot settings. CLM inherits these retrieval primitives but treats them as one component in a larger governed mesh rather than as the primary grounding mechanism. The critical departure is ontological: in CLM, retrieved content is anchored to an enterprise knowledge graph that constrains which entities, relations, and policies are admissible - a constraint that no retrieval mechanism alone can enforce.

### 3.3 Knowledge Graphs and Ontological Grounding

Hogan et al. [19] provide the canonical reference for knowledge graphs as a field, establishing the data models, query languages, and reasoning frameworks that underpin enterprise-scale graph deployments. GraphRAG [1] builds an entity graph from a private corpus and pre-computes community summaries, demonstrating that graph-augmented retrieval outperforms vector-only RAG on global sensemaking questions. A growing body of work explores LLM–knowledge graph synergies [6, 8] and applies ontological reasoning to fine-tune domain-specific enterprise language models [7]. Most recently, ontology-constrained neural reasoning architectures for enterprise agentic systems have argued that structured ontologies can constrain not only retrieval but also the reasoning surface of the agent, with controlled experiments reporting gains in accuracy, regulatory compliance, and role consistency for ontology-coupled agents over ungrounded baselines [9]. CLM takes this last observation as a first-class architectural decision: the corporate knowledge graph is not an accessory to the generative model - it is the grounding structure that makes generation corporate.

Sandkuhl et al. [26] extend this direction by proposing a meta-model that integrates prompt engineering with enterprise modeling notation, demonstrating through a systematic literature review of 21 papers that no existing approach addresses multi-perspective, integrated meta-models for LLM-guided enterprise modeling. CLM's relationship to this work is complementary rather than competitive: Sandkuhl et al. formalize how LLMs can be used within established enterprise modeling methods; CLM proposes the organizational knowledge foundation on top of which such methods - and the models that support them - are composed.

### 3.4 Neurosymbolic AI

The Neurosymbolic Mesh pillar of CLM draws explicitly on the neurosymbolic computing tradition. Garcez and Lamb [20] define neurosymbolic AI as the integration of neural learning with symbolic reasoning and explainability, arguing that neither paradigm alone is sufficient for systems that must be both flexible and trustworthy. The coupling CLM proposes - generative language models for fluency and contextual interpretation, corporate knowledge graphs for typed structure and governance - is a direct instantiation of this principle in an enterprise domain. The distinction from prior neurosymbolic architectures is scope: CLM applies neurosymbolic coupling not at the level of a single model but across an entire organizational knowledge substrate, with governance and sovereignty as non-negotiable properties of the coupling.

### 3.5 Small and Locally Deployable Language Models

The Phi-3 family [5] demonstrated that carefully curated training data yields compact, locally deployable models whose performance competes with much larger frontier systems. This is architecturally decisive for CLM: sovereign corporate deployment is technically viable when capable models can operate behind the firm's perimeter, eliminating the tension between model capability and data sovereignty. CLM composes task-specific Small Language Models alongside larger frontier models for tasks that demand broader reasoning, treating local deployability as a design floor rather than a deployment option.

### 3.6 Agentic Systems and Model-Context Protocols

Wang et al. [16] survey the current landscape of LLM-based autonomous agents, identifying auditability as a first-order unsolved problem across the field. The emergence of open tool-invocation protocols, and in particular the Model Context Protocol [10], provides a standardized way to expose corporate systems to language models as governed capabilities. CLM relies on this substrate: legacy systems are preserved as sources of truth and exposed as governed tools rather than replaced. The architectural commitment here is explicit - rip-and-replace is not the strategy - and it matters both for organizational risk tolerance and for adoption velocity.

### 3.7 Digital Twins as Functional Reasoning Surrogates

The Living Digital Twins construct in CLM draws on, and departs from, the established digital twin literature. Tao et al. [21] proposed the digital twin as a bidirectional coupling between a physical entity and its virtual model, driven by real-time data exchange; subsequent work extended this to product design frameworks [22] and smart manufacturing systems. These physical-world twins model artifacts and processes. CLM's Living Digital Twins model functional roles and reasoning patterns within an organization - the commercial twin, the financial twin, the operational twin - rather than physical entities. The inheritance is the surrogate-reasoning principle; the departure is the domain of application, which shifts from physical-process simulation to organizational-decision simulation, grounded in the corporate knowledge graph rather than in sensor data.

### 3.8 Explainability and Governed AI

The Governance Plane and the explainability-by-construction property of the Skill Graph connect CLM to the broader XAI literature. The NIST AI Risk Management Framework [12] and the OWASP Top 10 for LLM Applications [11] establish auditability and output traceability as baseline requirements for responsible AI deployment in regulated domains. CLM's contribution to this stream is architectural rather than methodological: explainability is not a post-hoc module applied to model outputs but a structural property of the Skill Graph - every recommendation is a traversal path whose nodes and edges constitute the explanation. This eliminates the need for approximation methods and makes explanation coextensive with the reasoning process itself.

### 3.9 Domain-Specific and Fine-Tuned Language Models

Recent deployments and an emerging analytical literature address the fine-tuning of foundation models on proprietary corpora as a path to domain specificity. Domain-Specific Language Models - fine-tuned or adapted foundation models whose weights encode the statistical regularities of a target domain - have been documented at enterprise scale by VM et al. [27], who demonstrate that parameter-efficient fine-tuning via QLoRA on domain-specific data yields locally deployable models competitive with larger frontier systems at a fraction of the memory cost. The category, variously labeled as DSLMs, domain-adapted LLMs, or enterprise-fine-tuned models, addresses P2 of the CLM design constraints - the absence of firm-specific priors in generic models - through training-time adaptation.
CLM departs from this approach in three respects that are architectural rather than parametric. First, fine-tuning encodes statistical regularities from a training corpus; it cannot capture the ongoing stream of tacit knowledge that accumulates through daily operations after the model is trained. CLM's Foundation Plane addresses P1 through continuous capture mechanisms that operate over live organizational interactions - a structural distinction from any training-time intervention. Second, fine-tuned models improve retrieval quality and domain fluency, but they do not produce compositional, auditable reasoning paths. CLM's Skill Graph addresses P3 by making the reasoning path the explanation, independently of the underlying model's weights. Third, sovereignty in fine-tuned deployments is a deployment option contingent on infrastructure choices; in CLM, sovereignty is an architectural floor - the system is designed from the outset so that data and weights do not leave the corporate perimeter by default. CLM is not an alternative to domain-specific fine-tuning: it is a foundation on which fine-tuned or locally deployed models are composed as one component among several.

### 3.10 Positioning Summary

Table 1 compares CLM to adjacent approaches across six dimensions that the five failure modes of Section 2 identify as non-negotiable. The comparison is conducted at the level of default architectural affordances - what each class of system provides by construction - not at the level of vendor-specific configurations. The six dimensions are: tacit-knowledge capture (whether undocumented reasoning is absorbed as a first-class asset); ontological grounding (whether outputs are bound to a typed enterprise ontology at generation time, not only at retrieval); explainability (whether the explanation is structural or reconstructed post hoc); data sovereignty (whether local-only operation is a precondition or an option); legacy integration (whether systems of record are governed sources of truth or are bypassed); and compositional reasoning (whether recommendations are composed over typed structures or sampled from a model distribution).
The comparison protocol matters because the most natural objection to CLM is that it is GraphRAG plus agent orchestration plus a compliance wrapper. It is not, and the distinction is architectural. CLM differs from GraphRAG because the knowledge graph is not only a retrieval scaffold but a co-active constraint on generation, governing which entities, relations, and policies are admissible in the output. It differs from platformized enterprise AI because governance and sovereignty are architectural floors that hold before any capability is exposed, rather than deployment options selected per installation. It differs from domain-specific fine-tuned models because tacit knowledge is captured continuously after training, not encoded statistically at training time.

**Table 1: Positioning of CLM against adjacent approaches**

| Dimension | Classical RAG | GraphRAG | Generic Agent Platforms | DSLM | Enterprise AI Platforms* | Corporate Language Model CLM |
|---|---|---|---|---|---|---|
| **Tacit knowledge capture** | No | No | No | No | Partial | **Yes** |
| **Ontological grounding** | Loose | Graph-level | No | Partial (statistical) | Varies | **Typed, co-active** |
| **Explainability** | Post-hoc | Partial | No | No | Varies | **By construction** |
| **Data sovereignty** | No | Optional | No | Optional | Optional | **Architectural floor** |
| **Legacy integration** | No | No | Via tools | No | Varies | **Governed, non-invasive** |
| **Compositional reasoning** | No | No | Emergent | No | No | **Typed Skill Graph** |

*Yes = native / by construction; No = not supported; Partial / Varies / Optional = capability present but inconsistent or configuration-dependent.*

**Enterprise AI Platforms denote vendor-integrated copilots and managed RAG suites (for example, patterns associated with Microsoft Copilot Studio, IBM watsonx, and ServiceNow AI); characterizations reflect general architectural patterns rather than specific product versions.*

And it differs from traditional digital twins because the modeled entity is organizational reasoning - how a function decides - rather than a physical system or process. Where these nine streams converge, CLM contributes a holistic architecture: a full-stack foundation where tacit capture, ontological grounding, hybrid retrieval, neurosymbolic coupling, twin-based reasoning, policy-aware execution, and sovereign deployment are co-designed from the outset rather than assembled as separate pipelines.

## 4 The CLM Concept: Five Capability Planes

CLM is organized as five interlocking capability planes. Four are stacked functional planes that follow the operational logic of enterprise knowledge work: what is known, how to act on it, how to deliver that action, and why the result can be trusted. The fifth plane, Deep Security, is transversal - it does not sit above or below the functional stack but wraps every plane end-to-end, enforcing sovereignty, auditability, and human oversight as conditions that hold before any capability is exposed rather than as properties added after the fact.

The ordering of the four functional planes is not arbitrary. Each plane presupposes the output of the one below it: the Tactical Plane cannot compose reasoning primitives without the grounded corporate memory that the Foundation Plane produces; the Wisdom-Out Plane cannot actuate without the composed reasoning that the Tactical Plane generates; the Governance Plane cannot audit what the Wisdom-Out Plane has not yet traced. This sequential dependency is the basis for the cumulative rollout model described in Section 7: stages cannot be inverted because planes cannot operate without their predecessors.

**Deep Security Layer (transversal)**

**Governance Plane** audit – ROI – rollback — Why we trust it

**Wisdom-Out Plane** actuation – evidence — How to deliver

**Tactical Plane** Skill Graph – Living Twins — Why we trust it

**Foundation Plane** ingestion – grounding - memory — What is known

### 4.1 Foundation Plane

The Foundation Plane is responsible for transforming organizational noise into queryable corporate memory. It ingests documents, conversations, emails, recorded calls, meeting transcripts, structured records, and multimodal artifacts; parses and cleans them; performs entity and relation extraction; and binds the resulting content to an enterprise ontology. Tacit-capture mechanisms operate continuously - distilling patterns of decision, exception, and rationale from everyday interactions rather than waiting for deliberate knowledge-documentation efforts. The output of this plane is not a collection of text chunks indexed into a vector store. It is a trainable, queryable, ontology-grounded corporate memory in which every element carries provenance, access permissions, and semantic typing. A vector store supports retrieval by similarity; the Foundation Plane's output supports retrieval by meaning, governance, and compositional reasoning.

The CLM Capability Architecture
Four Sequentially Dependent Planes under a Transversal Security Condition
Functional plane
Transversal security condition
Output required by the plane above
Deep Security Layer
Sovereignty · tenant isolation · rollback — a transversal condition of execution wrapping every plane
L04
Governance Plane
Logical traceability, audit, human-in-the-loop control, measurable ROI
audits what the plane below has traced
traced actuation
L03
Wisdom-Out Plane
Living Digital Twins and subagents acting in real time; outcome learning loop
actuates the reasoning composed below
composed reasoning
L02
Tactical Plane
Skill Graph — typed, composable, explainable operational reasoning
composes over the memory below
grounded corporate memory
L01
Foundation Plane
Tacit knowledge and legacy data transformed into ontologically anchored corporate memory
ingests documents, conversations, calls, transcripts, systems of record

Figure 1: The CLM capability architecture. The Deep Security Layer (outer frame) wraps four stacked functional planes as a transversal condition of execution. Planes are sequentially dependent: no plane can operate without the output of the plane below.

### 4.2 Tactical Plane

The Tactical Plane houses the reasoning primitives that the system applies to grounded context. It contains reusable skills - typed tactics, heuristics, and objection-handling patterns - typed personas, typed goals, and the Living Digital Twins that represent functional areas of the organization. A decision engine composes these primitives against the current context at runtime, selecting and combining the elements whose typed edges best match the situation. One architectural rule admits no exception: nothing in the Tactical Plane operates in the vacuum of a free-form prompt. Every reasoning step is anchored in a typed object that the system can inspect, audit, and evolve. This constraint is what makes the Tactical Plane's outputs explainable by construction rather than by approximation.

### 4.3 Wisdom-Out Plane

The Wisdom-Out Plane translates the Tactical Plane's composed reasoning into real-world actuation. Three transformations occur at this layer. First, the recommendation is rendered through the linguistic and decision-making style of the intended persona - the same underlying recommendation surfaces differently for a CFO than for a field sales representative, without the underlying logic changing. Second, the recommendation is routed to the correct stage of the business process, carrying evidence and reference cases that substantiate it. Third, the recommendation invokes governed tools against legacy systems of record through open protocols rather than bypassing or replacing those systems. This is the plane where what the system knows becomes what the organization does.

### 4.4 Governance Plane

The Governance Plane enforces what makes the system usable in regulated and high-stakes contexts. Every recommendation carries a traceable logical path from Skill to Objection to Persona to Goal - the typed edges of the Skill Graph described in Section 5.2 - that constitutes the explanation without requiring a separate explainability module. Measurable outcomes feed back into effectiveness scoring through an Outcome Learning Loop, updating skill weights based on observed results. Rollback is a first-class citizen of this plane: every governed action carries an explicit pause condition, a rollback procedure, and a defined fallback to human operation. The Governance Plane does not audit output after the fact; it structures every action so that audit is a property of the action's form, not a retrospective analysis applied to it.

### 4.5 Deep Security Layer

The fifth plane is transversal by design. Rather than treating security as a perimeter around the system, CLM models the Deep Security Layer as a set of concentric governance rings around the execution core. The generative engine operates at the center but reaches the real world only through successive checkpoints: a Governance Ring that enforces auditability and input sanitation; a Human-in-the-Loop Ring that inserts mandatory approval gates for high-impact operations; and an outer Sovereignty Ring that enforces tenant isolation, safe legacy write-back, and rollback plans. Data sovereignty, prompt-injection resistance, audit immutability, and human-in-the-loop gates are conditions that must hold before any capability is exposed - not features that can be deferred to a later release. The architectural and operational instantiation of this layer is expanded in Section 5.4.

The five planes are interlocking in a specific sense: they share a common typed object model. Skills, personas, objections, goals, evidence types, and governance policies are typed objects that all five planes can inspect and reason about. This shared typing is what makes the architecture composable rather than merely modular.

---

## 5 Architectural Pillars

Four pillars give the five capability planes their operational form. Each pillar addresses a distinct architectural problem: how to couple neural fluency with organizational structure (the Neurosymbolic Mesh); how to make tactical reasoning typed, composable, and explainable (the Skill Graph); how to model functional areas as reasoning surrogates (Living Digital Twins); and how to enforce sovereignty, traceability, and human oversight as non-negotiable properties of every execution path (the Deep Security Layer). The pillars operate as a coupled system: the Skill Graph traversal depends on the knowledge graph that the Neurosymbolic Mesh maintains; the twins hydrate context from that same graph; the Deep Security Layer wraps every interaction that any pillar exposes.

**Language Models**
(LLMs / SLMs)

Fluid generation
Context integration
Creative coverage

+

**Corporate Knowledge Graph**

Typed entities & relations
Business ontologies
Governance constrains

=

**Neurosymbolic Mesh**

Tacit knowledge
Made computable
Fluid outside, governed inside

The Four Architectural Pillars of CLM

*A Coupled System of Expressive Mechanisms under a Transversal Security Condition*

Architectural pillar | Transversal security condition | Coupling dependency

**Pillar IV · Deep Security Layer**

Sovereignty, traceability, and human oversight enforced as architectural floors — wraps every interaction that any pillar exposes

PILLAR I · GUARDED GENERATION

**Neurosymbolic Mesh**

Couples neural fluency with organizational structure: generative models and the enterprise knowledge graph operate co-actively at generation time, with symbolic validation of entities, relations, and policy before any output is emitted.

maintains the knowledge graph that traversal depends on

twins hydrate context from the same graph

PILLAR II · TYPED TRAVERSAL

**Skill Graph**

Makes tactical reasoning typed, composable, and explainable: skills, personas, objections, and goals as nodes of a typed directed graph; the reasoning path is the explanation.

PILLAR III · SURROGATE REASONING

**Living Digital Twins**

Models functional areas as reasoning surrogates that carry incentive structures and decision patterns, updated as the knowledge graph accumulates evidence.

Figure 2: The four architectural pillars of CLM. Each pillar operates across one or more capability planes and provides its distinctive expressive mechanism.

### 5.1 The Neurosymbolic Mesh

CLM rejects the binary choice between purely neural systems - fluid, contextually rich, but prone to hallucination and structurally opaque - and purely symbolic systems - rigorous and auditable, but brittle in the face of context and unable to generalize. This dichotomy, which Garcez and Lamb [20] identify as the central tension that neurosymbolic AI is designed to resolve, maps directly onto the enterprise AI problem: organizations need systems that are linguistically fluent at the interaction surface and governance-compliant at the decision surface.

CLM proposes instead a mesh in which three elements operate in concert. A generative substrate - composed of Large and Small Language Models selected by task - supplies fluency, contextual interpretation, and generalization. A corporate knowledge graph supplies typed entities, typed relations, explicit ontologies, business rules, and mathematical constraints that reflect the firm's actual operational reality. The coupling between the two is where the architectural contribution resides: rather than treating the knowledge graph as a retrieval index that feeds text to the model, the mesh makes the graph co-active at generation time. It constrains which entities can appear in outputs, which relations between them are valid under the current tenant's policy, which decision paths are admissible given the active governance context, and which claims require ontological validation before they reach the user.

**Guarded generation, not free decoding.** The phrase "co-active at generation time" invites a precise mechanical reading, and we give one. CLM does not claim universal constrained decoding; it implements ontological restriction through a hybrid guarded generation pipeline. For high-stakes tasks, the generative model does not directly emit unrestricted end-user output. It first emits a schema-bounded intermediate representation aligned to ontology-derived JSON Schemas or JSON-LD frames. A symbolic validator then checks entity admissibility, relation legality, policy constraints, and required provenance fields against the enterprise ontology and the active governance rules. Outputs that fail validation are not merely down-ranked; they are rejected and either deterministically repaired or re-generated under explicit violation feedback. Only validated artifacts are rendered into final user-facing language or executable specifications. In this sense, the knowledge graph constrains generation both ex ante, by bounding the admissible intermediate structures, and ex post, by validating every candidate artifact before release. This is stronger than purely post-hoc validation, because the generator is already channeled through typed structures, and safer than asserting full constrained decoding everywhere; it also aligns directly with the Spec-as-Code compiler of Section 6.

This guarded coupling is the reason CLM is not a variant of RAG. In a classical RAG pipeline, retrieved text is an input to the model - the model generates freely once retrieval is complete. In the Neurosymbolic Mesh, the knowledge graph participates in generation as a constraint: outputs that violate typed relations, exceed policy boundaries, or reference entities not present in the corporate ontology are structurally inadmissible rather than merely unlikely. This design choice is consistent with recent evidence that ontology-coupled enterprise agents outperform ungrounded agents on accuracy, regulatory compliance, and role consistency, with the largest gains where the model's parametric coverage is weakest [9], and with survey-level analyses reporting that knowledge-graph integration improves factual grounding and reduces hallucination in LLM systems [6, 8].

### 5.2 The Skill Graph

The Skill Graph is the most distinctive expressive construct in CLM and the primary mechanism by which the Tactical Plane achieves composable, explainable reasoning. A skill, in CLM terminology, is a typed, reusable unit of tactical reasoning: an objection-handling pattern, a persona-translation rule, a negotiation move, an expansion tactic, a discovery framework. Each skill is a first-class node in a directed graph, carrying a declared purpose, applicability conditions, composition rules, evidence requirements, risk level, and compliance constraints. Skills are not authored as free-text prompts. They are specified as structured objects with explicit activation signals, applicable personas, funnel stages, execution frameworks, and success metrics.

**A compact formalization.** The Skill Graph may be represented as a typed directed graph

$$G = (V, E, \tau_V, \tau_E) \qquad \text{(Eq. 1)}$$

where the vertex set partitions into typed subsets

$$V = V_s \cup V_p \cup V_o \cup V_g \cup V_f \qquad \text{(Eq. 2)}$$

with skill nodes $V_s$, persona nodes $V_p$, objection nodes $V_o$, goal nodes $V_g$, and funnel-stage nodes $V_f$; the node-typing function $\tau_V : V \rightarrow T_V$ assigns each node its type, where $T_V$ is the set of these five node types. Let $E \subseteq V \times V$ be a set of directed edges with edge-typing function $\tau_E : E \rightarrow R$, where the relation set

$$R = \{treats_objection,\ best_for_persona,\ supports_goal,\ complements_skill,\ works_in_stage,\ uses_evidence_type,\ enables,\ targets,\ primary_skill\} \qquad \text{(Eq. 3)}$$

This typing is declared in code rather than learned, which preserves graph traversability for explainability and ensures that the graph rebuilds deterministically from canonical registries - a property that makes the system's reasoning auditable at the structural level rather than dependent on stochastic model state.

Runtime selection is then a constrained subgraph-extraction problem. Given a context vector $c$ - the detected scenario, the active persona, the current funnel stage, and the stated or inferred goal - the decision engine first extracts the admissible candidate set $S(c) \subseteq V_s$ of skills whose typed edges are compatible with $c$, then ranks the candidates by

$$Score(s|c) = 0.4 \cdot A(s,c) + 0.4 \cdot E(s,c) - 0.2 \cdot R(s,c) \qquad \text{(Eq. 4)}$$

where $A(s, c)$ is adherence to the current context, $E(s, c)$ is observed empirical effectiveness, and $R(s, c)$ is execution risk encoded as a discrete coefficient (the coefficient values used in the current implementation are reported in Section 7.2); the constant 0.2 is the weight on the risk term, not a risk value. The output is not a single tactic but a coherent composition of complementary skills whose mutual consistency is guaranteed by the graph's typed edges. This formal core is deliberately minimal: it is sufficient to make the construct mathematically inspectable, while a fuller algebra of composition is left as future work (Section 9, RA2).

**Traversal Example**

Consider a B2B commercial scenario in which a CFO persona raises a surface price objection late in a negotiation. The decision engine detects four signals: *objection_type = price_surface*, *persona = CFO*, *stage = late_negotiation*, *goal = margin_preservation*. Traversal of the Skill Graph proceeds as follows: the engine identifies skills connected by *treats_objection → price_surface*, filters by *best_for_persona → CFO*, retains those that *works_in_stage → late_negotiation*, and ranks the resulting admissible set by Equation (4). The traversal returns a composition of three skills: a value-reframing tactic, a concession-boundary tactic, and a social-proof tactic. Each skill in the composition carries its own evidence requirements and compliance constraints, which the Wisdom-Out Plane enforces before actuation. The explanation for this recommendation is the traversal path itself - no post-hoc approximation is required, because the reasoning is the path.

The Skill Graph yields three properties that are structurally unavailable from a generic LLM. First, intelligent composition: the output is a coherent set of mutually consistent tactics, and the consistency is enforced by edge typing rather than by model judgment. Second, explainability by construction: every recommendation can be audited as a path through the graph without requiring a separate explainability module. Third, gap discovery: regions of the graph with no skills attached reveal blind spots in the organization's tactical coverage - personas, stages, or objection types for which no playbook has been developed.

### 5.3 Living Digital Twins

A Living Digital Twin is a computational surrogate for a functional area of the firm. CLM defines twins for the commercial function, the financial function, operations, customer success, and the owner or founder role. Each twin carries the incentive structure, decision patterns, preferred reasoning style, and linguistic register typical of its functional role - not as a static persona definition but as a set of typed parameters that the system updates as the corporate knowledge graph accumulates evidence about how that function actually operates.

The Living Digital Twin construct shares the surrogate-reasoning principle of the digital twin literature [21, 22] but departs from it in domain and grounding mechanism. Physical-world digital twins model artifacts and processes, grounded in sensor data and physical simulation. CLM's Living Digital Twins model organizational reasoning patterns, grounded in the corporate knowledge graph. Where a manufacturing digital twin simulates how a machine behaves under load, a CLM twin simulates how a CFO reasons under margin pressure - not by role-playing a persona, but by traversing the graph with the incentive weights and evidence thresholds that the financial function's accumulated decisions have established.

Rather than a metaphor, a twin is best understood as a computational abstraction with a fixed set of inspectable capacities. We decompose it into five capability bundles:

- **Capture capacity:** it absorbs and updates domain-specific reasoning traces from governed interactions, so that the twin's parameters track how the function actually decides rather than how it is nominally described.
- **Tactical capacity:** it selects and composes skills from the Skill Graph under contextual and policy constraints, anticipating second-order consequences before actuation.
- **Wisdom-Out capacity:** it renders function-specific recommendations or artifacts in the appropriate linguistic register without altering the underlying logical content.
- **Governance capacity:** it exposes provenance, confidence, approval thresholds, and rollback conditions for every recommendation it issues.
- **Security capacity:** it enforces access scope, tenant isolation, and policy conformance at the retrieval layer before any actuation is permitted.

At runtime, these capacities are exercised as an ordered sequence. A twin first hydrates context, retrieving from corporate memory only what is relevant, permitted, and current for the decision at hand - enforcing access control at the retrieval layer rather than at the output layer. It then simulates impact by traversing the Skill Graph to anticipate second-order consequences. It re-weights biases by reconciling competing functional perspectives through explicit ontological harmonization. Finally, it adjusts output by rendering the recommendation in the register appropriate to the intended audience.

Twins are not chatbots in costume. They are typed reasoning surrogates whose behavior is governed by the Skill Graph, whose context is bounded by the Governance Plane's access control, and whose outputs are always traceable to the corporate knowledge graph entries that produced them. More than a modeling convenience, this is where the paper's central claim becomes concrete: the firm's durable executable asset is not the generic model but the governed tacit-knowledge substrate that these capacities continuously cultivate and operationalize.

### 5.4 The Deep Security Layer

The security capacity that each twin exposes is enforced not by the twin itself but by the fourth pillar. The Deep Security Layer operationalizes the principle established in Section 4: sovereignty, auditability, and human oversight are conditions that hold before any capability is exposed, not properties added after deployment. It is modeled as three concentric governance rings around the execution core.

The Governance Ring is the innermost layer around the execution core. It enforces output auditability - every generated artifact carries a traceable provenance path - performs secret scanning to prevent sensitive data from appearing in outputs, and maintains a prompt-injection firewall that validates inputs against the corporate ontology before they reach the generative substrate. This ring operationalizes the OWASP Top 10 for LLM Applications [11] at the architectural level rather than as a post-deployment patch.

The Human-in-the-Loop Ring inserts mandatory approval gates for operations above a defined impact threshold. Any action that writes to a legacy system of record, commits a financial obligation, or modifies a customer-facing contract requires explicit human approval before execution. These approval decisions are themselves logged and feed back into the Governance Plane's effectiveness scoring, creating a closed loop between human judgment and system learning.

The Sovereignty Ring is the outermost layer and the architectural precondition for deployment in regulated industries. Corporate data - and, where technically feasible, the models themselves - remain inside the corporate perimeter, enabled by the maturation of compact, locally deployable language models [5]. Legacy systems of record are reached through governed tool invocation using open protocols such as the Model Context Protocol [10], preserving them as sources of truth rather than replacing them. Rollback plans are declared at the sovereignty layer: every governed action specifies its own rollback procedure and fallback to human operation before it is permitted to execute.

Four design commitments follow. Sovereignty is a floor, not a ceiling: data and models never leave the corporate perimeter by default, and exceptions require explicit governance override. Legacy systems are preserved, not replaced. Ground truth precedes generation: nothing is generated against sensitive operational contexts without passing ontological and policy validation first [9]. Human oversight is non-negotiable: the architecture does not assume that autonomous execution will be permitted at scale in regulated environments.

## 6 From Knowledge to Actuation: the Spec-as-Code Paradigm

A recurring failure mode of enterprise AI pilots is what we term the *semantic fall off the cliff*: the model produces a well-grounded recommendation, and the organization has no structural path to turn that recommendation into executed action without a separate, manual, error-prone engineering cycle. The gap is not incidental. It reflects a deeper architectural assumption embedded in most AI deployment models - that the system's responsibility ends at the recommendation boundary, and that translation into executable artifacts is someone else's problem.

In CLM, the Spec-as-Code paradigm is the software-engineering materialization of what can be operationally described as spec-driven corporate coding: business-authored, ontologically typed specifications become the canonical source from which downstream technical artifacts are compiled under ontological and governance constraints. Given a properly grounded intent, the system does not emit a summary or a recommendation. It compiles a structured set of four mutually consistent artifacts that close the loop between intent and executable action. The compilation draws on the full substrate of the Neurosymbolic Mesh - the grounded intent is validated against the corporate ontology, typed against the Skill Graph, and rendered through the Governance Plane before any artifact is produced.

A Business Specification captures the grounded intent in a schema-validated form: scope, stakeholders, acceptance criteria, measurable outcomes, governance constraints, rollback plan, and deployment strategy; every field is typed and validated against the corporate ontology. A Technical Specification translates the business intent into architecture decisions, system dependencies, and integration contracts. Behavior Scenarios are automatically derived from the acceptance criteria and expressed in BDD form - *Given the context, When this action occurs, Then this outcome is produced* - enabling regression testing from the first day of development. Interface Contracts specify the governed tools, data shapes, and exchange protocols at each integration point.

To make the pipeline concrete, the following illustrative example shows the typed entry point. A SemanticIntent is expressed and grounded by ontologies using JSON-LD, fixing the formal context from the origin (values below are drawn from the healthcare deployment of Section 7.3):

```
{
 "@context":
```

```
"https://example.org/clm/context.jsonld",
 "@type": "SemanticIntent",
 "intentId": "intent-discharge-summary-001",
 "domain": "clinical-documentation",
 "goal": "generate-jci-aligned-discharge-summary",
 "stakeholders": ["QualityCommittee", "ClinicalDocumentationTeam"],
 "constraints": {
  "regulatoryFramework": ["LGPD", "JCI"],
  "deploymentMode": "on-premise",
  "requiresHumanApproval": true
 },
 "acceptanceCriteria": [
  "Include mandatory JCI sections",
  "Preserve provenance for all extracted fields",
  "Reject output if ontology validation fails"
 ]
}
```

The compiler derives a typed Technical Specification from this intent. The excerpt below shows the validated derivation, with the validation stages and the integration contract that the Governance Plane will enforce:

```
{
 "techSpecId": "tspec-discharge-summary-001",
 "inputSchema": "ClinicalEncounter.v3",
 "validation": ["ontology-check", "policy-check", "provenance-check"],
 "integrationContract": {
  "sourceSystem": "EHR",
  "writeBack": false,
  "auditLog": "immutable"
 },
 "derivedFrom": "intent-discharge-summary-001"
}
```

The four artifacts are mutually consistent by construction, not by convention. The acceptance criteria in the Business Specification generate the BDD scenarios through a deterministic derivation; the scenarios bind the tests; the tests reference the Interface Contracts; the Interface Contracts constrain the Technical Specification. This chain of derivation has two operational consequences: change-impact analysis becomes tractable, and rollback stops being improvisational - the rollback plan is a typed field in the Business Specification, available to the Governance Plane as a structured artifact.

A property of the Spec-as-Code paradigm that deserves explicit attention is who drives the specification. We frame it not as a rhetorical inversion of authorship but as a disciplined, controlled reallocation of specification authority: domain experts author semantically typed requirements, and the system compiles them into downstream artifacts under ontological and governance constraints. In conventional AI-assisted development, the engineering team translates business intent into technical artifacts. In CLM's Spec-as-Code paradigm, the business team - or, as in the healthcare deployment described in Section 7.3, the customer's quality committee - drives the structural decisions in the Business Specification, and the CLM compiles those decisions into the downstream technical artifacts. This reallocation eliminates translation loss and makes the system's outputs auditable by domain experts who understand the business context.

The compilation process draws on three substrate layers operating in sequence. The grounded intent is first validated against the corporate ontology in the Foundation Plane, ensuring that scope, stakeholders, and acceptance criteria are typed against entities and relations that exist in the firm's knowledge graph. The validated intent is then resolved through the Skill Graph in the Tactical Plane, which maps acceptance criteria to applicable skills, risk levels, and evidence requirements. Finally, the Governance Plane enforces policy constraints and rollback conditions before any artifact is emitted. The four output artifacts are mutually consistent by construction because they share a single derivation chain: the acceptance criteria in the Business Specification generate the BDD scenarios deterministically, the scenarios bind the test suite, and the test suite references the Interface Contracts. Specific scoring functions, registry data structures, and compilation heuristics constitute proprietary implementation details that are outside the scope of this paper.

---

## 7 Early Empirical Evidence

This paper is, by design, a position paper. The evidence presented in this section does not validate the CLM architecture in the sense of a controlled empirical study; it demonstrates that the architecture is instantiable, that its constraints are operationally enforceable in a regulated environment, and that the cumulative-stage model it proposes is consistent with observed deployment sequencing. We present three categories of evidence - structural, operational, and convergent - and conclude with an explicit enumeration of what has not yet been measured.

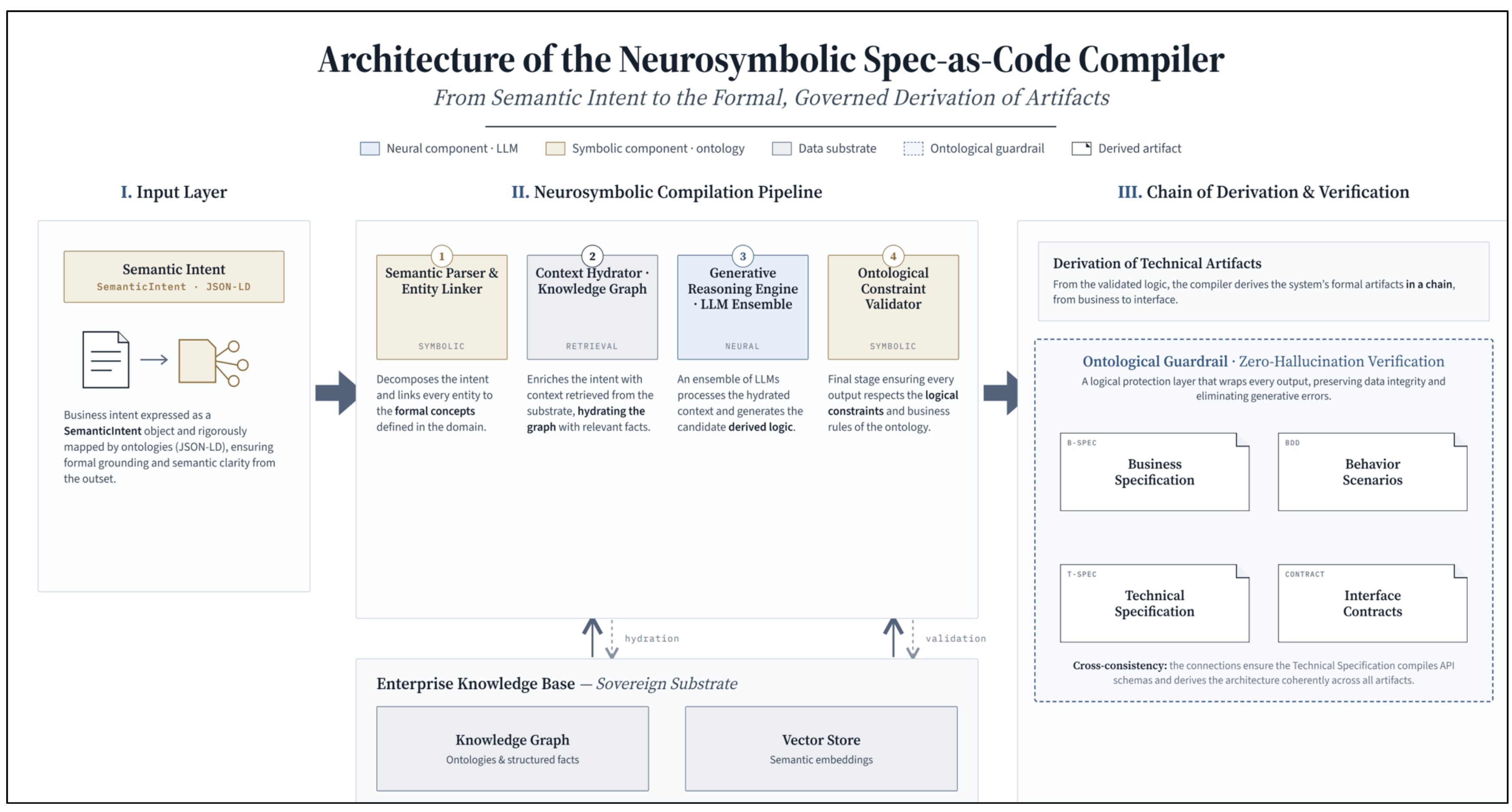

Figure 3: Architecture of the Neurosymbolic Spec-as-Code Compiler. The workflow moves from a business intent expressed as a SemanticIntent (Input Layer), through a four-stage hybrid pipeline - Semantic Parser and Entity Linker, Knowledge Graph Context Hydrator, Generative Reasoning Engine, Ontological Constraint Validator - to a chain of derived artifacts (B-SPEC, BDD, T-SPEC, Interface Contracts) produced under an Ontological Guardrail that enforces cross-artifact consistency.

### 7.1 Cumulative Maturity Model

CLM is a substrate, not a product, and organizations adopt it cumulatively. We propose six stages of capability maturation, each of which presupposes the output of the preceding stage and unlocks the next.

- **Stage 1 - Capture and Grounding.** Ingestion pipelines, cleaning, entity extraction, and ontology binding are operational. The knowledge base is live and queryable. This stage is the precondition for all subsequent stages.
- **Stage 2 - Reliable Retrieval.** Hybrid retrieval - combining dense vector search, lexical signals, and graph traversal - is operational with ontological grounding guarantees.
- **Stage 3 - Twinization.** Living Digital Twins are activated for the first critical functional areas - typically commercial, financial, and operational. Twin session state is persistent and access-controlled by department tier.
- **Stage 4 - Specialized Models.** Domain-specific Small Language Models are fine-tuned or adapted - through supervised fine-tuning (SFT) or direct preference optimization (DPO) - for tasks where generic frontier models underperform.
- **Stage 5 - Spec-as-Code.** The compilation from grounded intent to executable artifacts is operational. Organizations operating at this stage begin to deliver specification-first rather than implementation-first.
- **Stage 6 - Governed Actuation.** Integration with legacy systems of record is mature. Governed write-backs, human approval gates, and rollback procedures are fully operational under audit.

Two properties of this model are architecturally non-negotiable. Each stage is valuable in itself; and no stage is skippable: attempting governed actuation (Stage 6) without twinization (Stage 3) or reliable retrieval (Stage 2) is the architectural root cause of most failed enterprise AI projects.

### 7.2 Structural Evidence: Care Engine Platform

The Care Engine Platform is Care Intelligence's production instantiation of the CLM framework: CLM designates the conceptual architecture described in Sections 4 through 6; Care Engine is the system that implements it.

*Scope: the evidence in this subsection is structural and parameter-level. It demonstrates that the CLM architecture exists as a running implementation with declared, inspectable parameters - not that it outperforms any baseline or generalizes across contexts. All figures are drawn from the implementation snapshot dated April 26, 2026.*

The CLM-specific test suite contains 90 dedicated test methods across five modules; the broader repository suite of 232 tests passes in approximately 24.5 seconds. The Outcome Learning Loop mechanism is fully tested, but no production signals have yet been persisted.

**Skill Graph Composition**

The first canonical instantiation of the CLM Skill Graph contains 41 typed nodes - 12 skills organized across five categories (Expansion, Persuasion, Translation, Objection-handling, and Closing), 5 personas, 5 playbooks, 5 sub-agents wired through skill registries, 2 objection patterns, 7 funnel stages, and 5 goal types - connected by 100 typed edges across nine relation types. Node and edge typing are declared in code rather than learned. The graph rebuilds deterministically from canonical registries on each process start.

**Decision Engine Determinism and Parameter Auditability**

The Decision Engine orchestrates skill activation using the explicit weighted formula of Equation (4), with risk levels mapped to fixed coefficients (low = 0.1, medium = 0.3, high = 0.6). These coefficients are the values that R(s, c) assumes; the constant 0.2 in Equation (4) is the weight on the risk term, a distinct quantity. Skill effectiveness is updated through an Exponential Moving Average ($\alpha = 0.3$) on outcome signals, with declared score deltas (success: +0.15; partial: +0.05; failure: −0.10; abandoned: −0.05) over a rolling window of the last 500 events. System behavior at the orchestration layer is auditable at the parameter level: every activation decision is traceable to declared weights rather than to opaque model state.

Three design choices in this parameterization merit brief explanation. The symmetric weights on adherence and effectiveness (0.4 each) reflect equal priority assigned to process compliance and outcome quality in the pilot deployment context; the risk penalty is set at 0.2 - half the weight of either primary criterion - to permit compositions that include higher-risk tactics when both adherence and effectiveness scores are high, avoiding a regime that structurally excludes bold approaches in early deployment. The EMA decay rate ($\alpha = 0.3$) places the system in a moderately responsive regime: each new outcome signal contributes 30% of the updated effectiveness estimate, with 70% flowing from prior evidence. This rate balances responsiveness to recent deployment conditions against resistance to signal noise from individual interaction outcomes. The score deltas are intentionally asymmetric - success increments (+0.15) exceed failure decrements (−0.10) in absolute value - creating a mild positive exploration bias that avoids penalizing skill discovery as harshly as confirmed failure.

**Stage Rollout, Observed**

Of the six cumulative stages defined in Section 7.1, Stages 1 through 3 are in production. Stage 4 is in test, with 12 canonical skills and 6 sub-agents constituting the Phase 1 minimum viable product. Stages 5 and 6 remain in design. No stage inversion was observed - Stages 1 through 3 reached production before Stages 4 through 6 were attempted. This observation is consistent with the cumulative-stage claim; it does not constitute validation of the claim, since the sample is a single organization.

**Pre-Production Limitations**

Three design-level limitations are acknowledged explicitly. The skill-context matcher uses case-insensitive substring matching on signal keywords, which can produce false positives. Twin session state is held in process memory and is lost on restart; durable persistence is not yet implemented. The Skill Graph silently returns an unavailable status if the NetworkX dependency is absent from the runtime rather than raising an explicit exception. None of these are operational incidents; they are pre-production constraints that scope the present evidence.

### 7.3 Operational Evidence: Healthcare Deployment

*Scope: the evidence in this subsection is operational. It demonstrates that the CLM architecture is deployable in a regulated environment under real sovereignty and auditability constraints - not that it produces outcomes superior to alternatives. No comparison against a baseline was performed.*

**Foundation Plane: Clinical Aggregator**

The first production deployment of the CLM Foundation Plane was implemented at a JCI-accredited tertiary teaching hospital ranked among the top ten in Brazil. Production go-live was scheduled for April 29, 2026, following completion of the homologation cycle. The deployed component - the Clinical Aggregator - performs agnostic capture from the hospital's electronic health record (EHR), parses and structures the captured content against a clinical ontology, and exposes it as a queryable substrate for downstream decision-support agents. The Clinical Aggregator was deployed as a containerized component running inside the hospital's EHR environment, jointly homologated by the hospital and the EHR provider, with no data exported beyond the hospital perimeter.

**Sovereignty Ring: Local-Only De-identification**

Upstream of the Clinical Aggregator, a local-only de-identification module operates entirely within the hospital perimeter. Personal identifying information is removed before any downstream processing, with no external retention at any stage. This module operationalizes the Sovereignty Ring described in Section 5.4 and was a precondition for hospital approval under Brazil's LGPD framework (Law 13,709/2018). Quantitative performance figures for the de-identification component are part of the doctoral research referenced in Section 7.4.

**Customer-Driven Spec-as-Code: JCI-Aligned Discharge Summary**

As part of the same engagement, the hospital's quality committee identified a deliverable required for renewal of its Joint Commission International accreditation: a discharge summary in JCI-aligned format. The specification was defined jointly with the hospital's quality team and compiled into the four artifact types described in Section 6. The relevant property of this case is the controlled reallocation of specification authority: the hospital's quality team (not the implementing organization's engineering team) drove the structural decisions, and the CLM compiled them into mutually consistent executable artifacts.

### 7.4 Convergent Research: Doctoral Validation

*Scope: the evidence in this subsection is convergent. An independent doctoral research line applies a graph-typed knowledge approach in the same clinical environment. Quantitative results are out of scope for this paper and will be reported in the doctoral dissertation.*

An applied doctoral research project at the Business Graduate Program at Unisinos University (qualified in January 2024, with defense scheduled by September 2026) applies a graph-typed knowledge approach to value-based healthcare measurement under the Porter [13] framework. The work has institutional research ethics approval and is conducted at the same hospital described in Section 7.3. The corporate-domain CLM framework presented here is the broader generalization of which the doctoral instantiation is a domain-specific case. The doctoral work is the source of the quantitative empirical evaluation that this paper consciously does not report; those results will be presented in the doctoral dissertation.

### 7.5 Limitations of the Present Evidence

The evidence in Sections 7.2 through 7.4 is constrained in ways that must remain explicit. No comparison against public baselines or competing architectures was performed. No formal ablation study isolates the contribution of individual pillars. The Outcome Learning Loop has not accumulated production signals. The temporal window is narrow: structural counts and parameters are from a single April 2026 snapshot. Accordingly, the present evidence supports feasibility, architectural coherence, and deployment viability under regulatory constraints, but it does not support claims of superiority, causality, or cross-organizational generalizability.

These constraints are intrinsic to early-stage architectural papers and do not refute the propositions of Sections 4 through 6. They delimit what the present evidence can and cannot establish: the architecture is instantiable, its sovereignty and auditability constraints are enforceable under real regulatory conditions, and its cumulative-stage model is consistent with observed deployment sequencing. Each constraint maps directly to a research direction identified in Section 9.

---

## 8 Key Differentiators and Strategic Properties

Five properties, taken together, distinguish CLM from both vanilla RAG systems and from the current generation of enterprise AI platforms. We present them not as empirically validated superiorities but as architectural commitments whose strategic implications are measurable in principle and whose design rationale has been established in Sections 4 through 6.

**Computable Tacit Knowledge**

CLM treats tacit-knowledge capture as a first-class architectural concern. The Foundation Plane continuously distills patterns of decision, exception, and rationale from everyday organizational interactions, converting what Nonaka and Takeuchi [17] identified as the primary source of organizational competitive advantage into a queryable, typed, ontologically grounded substrate. This produces a compounding property we term the *Wisdom Listener effect*[1]: each interaction extends the captured tacit substrate available for subsequent reasoning, creating a self-reinforcing loop between system usage and system competence.

This compounding property bears structural similarity to two established mechanisms. Teece [23] identifies orchestration - the capacity to continuously sense, recombine, and renew knowledge assets - as the microfoundation of dynamic capabilities that sustains competitive advantage over time; the Wisdom Listener effect is the system-level, continuous analogue of that capacity, replacing episodic managerial judgment with an architectural feedback loop. Argote and Miron-Spektor [24] document organizational learning curves in which accumulated experience produces non-linear improvements in performance; the Foundation Plane operationalizes this accumulation computationally, making the learning curve a structural property of the architecture rather than a function of workforce tenure. The distinction from both precedents is mechanism: CLM grounds the compounding loop in a typed, governed substrate rather than in social interaction or human memory - which makes the effect transferable across personnel transitions and auditable at the organizational level.

**Compositional Reasoning via the Skill Graph**

Recommendations produced by CLM are not sampled from a model's probability distribution. They are composed from a typed graph whose edges encode tested combinations of tactics, personas, objections, and goals. The reasoning process is closer to operations research - constraint satisfaction over a typed graph - than to prompt engineering over a generative model. For organizations operating in regulated or high-stakes contexts, this is the difference between a recommendation whose basis can be reconstructed and one that cannot.

**Living Twins as Organizational Reasoning Surrogates**

Each Living Digital Twin is a typed reasoning surrogate grounded in the corporate knowledge graph, not a persona-skinned chatbot. The distinction matters because a surrogate reasons from the functional area's actual incentive structure, evidence thresholds, and decision patterns - derived from accumulated organizational interactions - rather than from a static role description. As the Foundation Plane accumulates tacit knowledge, the twins become more accurate representations of how each functional area actually decides, not how it is supposed to decide.

### Sovereignty and Auditability as Architectural Floors

In CLM, sovereignty is an architectural floor: the system is designed so that data and models never leave the corporate perimeter by default, and the capability to operate locally is a precondition for the architecture rather than an optional deployment mode. Auditability follows the same logic: every recommendation carries a traceable logical path that is a structural property of the Skill Graph traversal, not a post-hoc explanation generated separately.

### Explainability by Construction

Because CLM's recommendations are paths through a typed graph, the explanation is coextensive with the reasoning: this skill, because of this objection, for this persona, supporting this goal, with this evidence type. No approximation method is required. The system's reasoning can be inspected, challenged, and improved by domain experts who understand the business context but not the underlying model - the population that matters most in organizational deployment.

These five properties compound in a specific sequence. As a firm accumulates tacit knowledge inside the Foundation Plane, the Tactical Plane becomes more discriminating; as skill selection improves, the twins become more accurate; as twin accuracy improves, output quality increases; as output quality increases, human approvers delegate more consequential work, which in turn enriches the tacit layer further. CLM is, by design, a compounding architecture.

---

## 9 Discussion and Research Agenda

This paper has made a conceptual argument: that the right unit of analysis for enterprise AI is a foundation rather than a model or a pipeline, and that treating sovereignty, auditability, and tacit capture as design constraints rather than as deferred features produces a categorically different class of system. The early evidence presented in Section 7 demonstrates instantiability under real regulatory conditions; it does not validate the architecture in the sense of a controlled comparative study. The six research directions below define the empirical and theoretical agenda that the architecture implies. They are presented as a numbered agenda to enable direct citation by work that builds on or challenges this framework.

**RA1 - Empirical validation at scale.** Comprehensive validation requires controlled comparison against RAG baselines, GraphRAG-style deployments, domain-specific fine-tuned models, and commercial enterprise AI platforms across multiple verticals. Of particular interest is whether the Wisdom Listener effect produces measurable compounding in retrieval quality and recommendation accuracy over deployment cycles - a longitudinal measurement program that the current evidence window cannot support. The doctoral research referenced in Section 7.4 will contribute the first quantitative before-and-after outcomes from a regulated deployment; cross-vertical replication remains as subsequent work.

**RA2 - Formal theory of the Skill Graph.** Section 5.2 gives the Skill Graph a compact formal core: a typed directed graph $G = (V, E, \tau_V, \tau_E)$ with runtime selection defined as constrained subgraph extraction under a context vector followed by ranking. A fuller treatment of its compositional properties - the conditions under which skill combinations produce coherent outputs, the algebraic structure of the traversal, and the completeness guarantees that can be made about gap discovery - remains to be developed. Such a formalization would connect the Skill Graph to the broader literature on typed-graph formalisms and constraint satisfaction.

**RA3 - Generalization across verticals.** The architectural argument is vertical-agnostic by design: the five failure modes of Section 2 recur across industries regardless of domain. The density of tacit knowledge, the regulatory envelope, and the risk tolerance of the domain are non-trivial moderators of how each capability plane is instantiated, however. The conditions under which architectural primitives generalize while the policy layer varies - and when domain adaptation requires architectural modification rather than parameterization - is an empirical question that cross-vertical deployment will answer. The convergence between CLM's design constraints and the findings of Görgen et al. [25] and Sandkuhl et al. [26] - each independently identifying ontological grounding as an open problem in enterprise AI contexts - suggests that the problem space CLM addresses is being recognized across research communities, strengthening the case that the architectural response proposed here addresses a structurally real gap rather than a locally perceived one.

**RA4 - Sovereignty-performance trade-off under local models.** CLM's sovereign deployment commitment relies on the continued maturation of compact, locally deployable language models [5]. The performance gap between locally deployed Small Language Models and frontier cloud-hosted models is narrowing but not eliminated. A systematic quantitative study of this trade-off - measuring recommendation quality, retrieval accuracy, and latency as a function of model size and deployment context - would provide the architectural guidance needed to operationalize the sovereignty commitment across the full range of enterprise environments.

**RA5 - Human and organizational conditions.** CLM is an architecture for an organization willing to treat its tacit knowledge as infrastructure worth building. The system presupposes organizational intent: a firm that deploys CLM as a faster chatbot will realize a fraction of its potential. The human and organizational conditions under which this intent is sustained - change management requirements, expertise needed to author and maintain the Skill Graph, governance structures for managing twin accuracy over time - open a research program connecting CLM to the organizational knowledge management literature that Section 3.1 identified as foundational.

**RA6 - Open protocol dependency.** The portability of CLM's governed-tool substrate depends on the continued maturation of open agent-to-system protocols, most prominently the Model Context Protocol [10]. Fragmentation of this ecosystem would partially reintroduce the integration complexity that CLM is designed to eliminate. Monitoring this dependency and developing fallback integration strategies is a near-term engineering concern with immediate operational consequences, independent of CLM's theoretical merits.

## 10 Conclusion

The central argument of this paper is a repositioning. Enterprise AI has been framed, for the past several years, as a model selection problem: which foundation model produces the best outputs? This framing is incomplete - not because model quality is irrelevant, but because it places the competitive question at the wrong level of the stack. A generic model, however capable, carries no priors about how a specific organization decides, negotiates, or executes. The knowledge that makes organizations effective - their tacit reasoning patterns, their accumulated exceptions, their governance constraints, their execution contracts - is not accessible to a model that has never encountered it. Selecting a better model does not solve this problem. Building a deeper foundation does.

We have specified that foundation as a Corporate Language Model - an enterprise knowledge foundation organized as five capability planes and four architectural pillars, with a Deep Security Layer that is transversal by design. CLM is not a replacement for a generic LLM. It is the structured layer that transforms general linguistic competence into corporate executable intelligence - the scaffolding around it, below it, and at times orthogonal to it that makes the difference between a system that answers questions and a system that executes decisions.

The strategic implication is direct and compounding. Firms that operate in the vacuum of generic models will increasingly compete against organizations where tacit knowledge has become their largest executable asset - where the foundation is deep enough that every interaction makes the system more accurate, every governed action makes the audit trail richer, and every deployment cycle makes the competitive gap wider. CLM makes the conceptual case that such organizations are not only possible but architecturally specifiable. The question is no longer whether enterprise AI can encode organizational knowledge. It is whether organizations are willing to treat that knowledge as infrastructure worth building.

Future work will present empirical evaluation in targeted verticals; formal analysis of the compositional properties of the Skill Graph; quantitative studies of the sovereignty-performance trade-off under locally deployed small models; and longitudinal measurement of the Wisdom Listener effect across deployment cycles. The doctoral research referenced in Section 7.4 will contribute the first quantitative before-and-after outcomes from a regulated healthcare deployment. Together, these lines of work will move CLM from a conceptual architecture to an empirically grounded framework - the progression that every serious position paper is designed to initiate.

**Acknowledgments**
The CLM framework was originally conceived and developed by Fabricio Colvero Avini, doctoral candidate at Unisinos University and founder of Care Intelligence; authorship is registered with the Câmara Brasileira do Livro (CBL), dated April 23, 2026. Guilherme Trez, Professor at Unisinos, joined Care Intelligence and contributed to the consolidation of the framework for academic publication; the doctoral research referenced in Section 7.4 is conducted at the same institution. The authors thank the Care Intelligence team for sustained implementation work that, while outside the scope of this paper, grounds the validity of the conceptual claims presented here, and the quality committee at the partner hospital whose collaborative engagement shaped the customer-driven specification described in Section 7.3.

**AI-Assisted Writing Disclosure**
In line with current best practice for transparency in scholarly writing [14, 15], the authors declare the use of generative AI assistance during the preparation of this manuscript. The conceptual architecture, all design decisions, the five capability planes, the four architectural pillars, the Skill Graph formalism, the Living Digital Twins construct, the Deep Security Layer, the Spec-as-Code paradigm, and the integration of these elements into a single framework were originally conceived by the lead author over an extended period of work and registered with the Câmara Brasileira do Livro on April 23, 2026. Generative AI tools (specifically, large language models from Anthropic) were used during the writing process for language refinement, structural organization of the manuscript, and editorial assistance, under continuous supervision and review by the authors. AI tools did not contribute to the intellectual substance of the work. All references in this paper were independently verified by the authors against their original sources. Responsibility for all claims, positions, and any errors that remain rests solely with the authors.

[1] Term coined in this paper to denote the property by which a tacit-capable foundation accumulates operational value through use: each interaction extends the captured tacit substrate available for subsequent reasoning, producing a self-reinforcing loop between system usage and system competence.